\documentclass[10pt,journal,compsoc]{IEEEtran}
\ifCLASSOPTIONcompsoc
  \usepackage[nocompress]{cite}
\else
  \usepackage{cite}
\fi
\usepackage{graphicx}
\usepackage{amsfonts}
\usepackage{xcolor}
\usepackage{wrapfig,lipsum,booktabs}

\newcommand{\etal}[1]{#1~\textit{et al.}}
\usepackage{amssymb}
\usepackage{pifont}
\newcommand{\cmark}{\ding{51}}%
\newcommand{\xmark}{\ding{55}}%

\usepackage{hyperref}

\ifCLASSINFOpdf
\else
\fi
\usepackage{bbm}

\newcommand{\newcolor}{\textcolor{black}}
\newcommand{\newcolortwo}{\textcolor{black}}
\newcommand{\newcolorthree}{\textcolor{black}}
\begin{document}
%
\title{Explainability in Graph Neural Networks:\\A Taxonomic Survey}
%
%
%
%

\author{Hao Yuan,
        Haiyang Yu, 
        Shurui Gui, 
        and Shuiwang Ji,~\IEEEmembership{Senior Member,~IEEE}
\IEEEcompsocitemizethanks{\IEEEcompsocthanksitem H. Yuan, H. Yu, S. Gui, and S. Ji are with Department of Computer Science and Engineering, Texas A\&M University, College Station, TX 77843, USA, e-mail: \{hao.yuan, haiyang, shurui.gui, sji\}@tamu.edu.
}
}

%
%

\markboth{Preprint}%
{Yuan \MakeLowercase{\textit{et al.}}: Bare Demo of IEEEtran.cls for Computer Society Journals}
%



\IEEEtitleabstractindextext{%
\begin{abstract}
Deep learning methods are achieving ever-increasing performance on many artificial intelligence tasks. A major limitation of deep models is that they are not amenable to interpretability. This limitation can be circumvented by developing post hoc techniques to explain predictions, giving rise to the area of explainability. Recently, explainability of deep models on images and texts has achieved significant progress. In the area of graph data, graph neural networks (GNNs) and their explainability are experiencing rapid developments. However, there is neither a unified treatment of GNN explainability methods, nor a standard benchmark and testbed for evaluations. In this survey, we provide a unified and taxonomic view of current GNN explainability methods. Our unified and taxonomic treatments of this subject shed lights on the commonalities and differences of existing methods and set the stage for further methodological developments. To facilitate evaluations, we provide a testbed for GNN explainability, including datasets, common algorithms and evaluation metrics. Furthermore, we conduct comprehensive experiments to compare and analyze the performance of many techniques. Altogether, this work provides a unified methodological treatment of GNN explainability and a standardized testbed for evaluations.
\end{abstract}

}

\maketitle

\IEEEdisplaynontitleabstractindextext

%
\IEEEpeerreviewmaketitle

\IEEEraisesectionheading{\section{Introduction}\label{sec:introduction}}

%
%
%
%


\IEEEPARstart{T}he developments of deep neural networks have revolutionized the fields of machine learning and artificial intelligence. Deep neural networks have achieved promising performance in many research tasks, such as computer vision~\cite{ji20133d, simonyan2014very}, natural language processing~\cite{vaswani2017attention}, and graph data analysis~\cite{kipf2016semi, xu2018powerful}. These facts motivate us to develop deep learning methods for real-world applications~\cite{wang2020moleculekit, MaoGeo-Spatiotemporal, Yuan:bioinfo18, maruhashi2018learning } in interdisciplinary domains, such as finance, biology, and agriculture etc. However, since most deep models are developed without interpretability, they are treated as black-boxes. Without reasoning the underlying mechanisms behind the predictions, deep models cannot be fully trusted, which prevents their use in critical applications pertaining to fairness, privacy, and safety. In order to safely and trustfully deploy deep models, it is necessary to provide both accurate predictions and human-intelligible explanations, especially for users in interdisciplinary domains. These facts raise the need of developing explanation techniques to explain deep neural networks.

The explanation techniques of deep models generally study the underlying relationships behind the predictions of deep models. Several techniques are proposed to explain the deep models for image and text data~\cite{simonyan2013deep, smilkov2017smoothgrad,zhou2016learning, selvaraju2017grad, dabkowski2017real, yuan2020interpreting,olah2017feature, olah2018the, yang2019xfake, du2018towards, wang2020icapsnets, Du:www19}. These methods can provide input-dependent explanations, such as studying the important scores for input features, or a high-level understanding of the general behaviors of deep models. For example, by studying the gradients or weights~\cite{simonyan2013deep, smilkov2017smoothgrad, yang2019xfake}, we can analyze the sensitivity between the input features and the predictions. Existing approaches~\cite{zhou2016learning, selvaraju2017grad, du2018towards} map hidden feature maps to the input space and highlight the important input features. In addition, by occluding different input features, we can observe and monitor the change of predictions to identify important features~\cite{dabkowski2017real, yuan2020interpreting}. Meanwhile, several studies~\cite{olah2017feature, simonyan2013deep} focus on providing input-independent explanations, such as studying the input patterns that maximize the predicted score of a certain class. Furthermore, the meaning of hidden neurons is explored to understand the whole prediction procedures~\cite{yuan2019Interpreting, olah2018the}. Recent survey works\cite{du2019techniques, rai2020explainable, dovsilovic2018explainable, molnar2019} provides good systematic reviews and taxonomies of these methods. However, these surveys only focus on the explanation methods for image and text domains and ignore the 
explainability of deep graph models.

Recently, Graph Neural Networks (GNNs) have become increasingly popular since many real-world data are represented as graphs, such as social networks, chemical molecules, and financial data~\cite{ma2020deep,book2, tolmachev2021bermuda}. Several graph-related tasks are widely studied, such as node classification~\cite{gao2019graph, henaff2015deep, Liu:Deeper}, graph classification~\cite{xu2018powerful, zhang2018end}, link predictions~\cite{zhang2018link,cai2020multi,cai2020line}. In addition, many advanced GNN operations are proposed to improve the performance, including graph convolution~\cite{kipf2016semi, liu2020non, Gao:KDD18}, graph attention~\cite{velivckovic2017graph, thekumparampil2018attention}, and graph pooling~\cite{Yuan2020StructPool, wang2020second, gao2020topology}. However, compared with image and text domains, the explainability of graph models are less explored, which is critical for understanding deep graph neural networks. Recently, several approaches are proposed to explain the predictions of GNNs~\cite{xie2022task}, such as XGNN~\cite{xgnn}, GNNExplainer~\cite{ying2019gnnexplainer}, PGExplainer~\cite{luoparameterized}, andSubgraphX~\cite{yuan2021explainability}, etc.  
These methods are developed from different angles and provide different levels of explanations. In addition, it is still lacking standard datasets and metrics for evaluating the explanation results. Hence, it raises the need of a systematic survey covering the methods and evaluations of GNN explanation techniques.

To this end, this survey provides a systematic study of different GNN explanation techniques. We aim at providing intuitive understanding and high-level insights of different methods so that researchers can choose proper directions to explore. The contributions of this work are summarized as below:
\begin{itemize}
    \item This survey provides a systematic and comprehensive review of existing explanation techniques for deep graph models. To the best of our knowledge, this is the first and only survey work on such a topic.
    \item We propose a novel taxonomy of existing GNN explanation techniques. We conclude the key idea of each category and provide insightful analysis.
    \item We introduce each GNN explanation method in detail, including its methodology, advantages, drawbacks, and difference compared with other methods. 
    \item We summarize the commonly employed datasets and evaluation metrics for GNN explanation tasks. We discuss their limitations and recommend several convincing metrics. 
    \item We build three human-understandable datasets from the text domain by converting sentences to graphs. These datasets are now publicly available and can be directly used for GNN explanation tasks. 
    \item We develop an open-source library for the GNN explanation research. In this library, we include the implementations of several existing techniques, commonly used datasets, and different evaluation metrics. It can be used to reproduce existing methods and develop new GNN explanation techniques. 
    \item \newcolor{We conduct experiments to study and compare several explanation techniques. Our analysis can provide insights for new researchers about the development of the research field and baseline selections.}
\end{itemize}

\noindent{{\textbf{Explainability versus Interpretability:} 
Explainable artificial intelligence is an emerging area of research, and the use of terminology is not completely standardized. In particular, the differences of the terms ``explainability'' and ``interpretability'' are not completely defined. In some studies, these two terms are used interchangeably~\cite{molnar2019}, while in other studies some subtle differences have been made. In this work, we argue that these two terms should be differentiated. In particular, we follow the existing work~\cite{rudin2019stop} to distinguish these two terms. We consider a model to be ``interpretable'' if the model itself can provide humanly understandable interpretations of its predictions. Note that such a model is no longer a black box to some extent. For example, a decision tree model is an ``interpretable'' one. 
Meanwhile, an ``explainable'' model implies that the model is still a black box whose predictions could potentially be understood by \emph{post hoc} explanation techniques. We articulate such a difference with the hope that the community could eventually develop a standardized use of terminology as this field matures.
}} 
\section{The Challenges}

Explaining deep graph models is an important but challenging task. In this section, we discuss the common challenges for explaining GNNs. 

First, unlike images and texts, graphs are not grid-like data, which means there is no locality information and each node has different numbers of neighbors. Instead, graphs contain important topology information and are represented as feature matrices and adjacency matrices. To explain feature importance, we may directly extend the explanation methods for image data to graph data. However, the adjacency matrices represent the topology information and only contain discrete values. \newcolor{Then existing methods may not be suitable for obtaining high-quality explanations for GNNs.} For example, input optimization methods~\cite{olah2017feature, simonyan2013deep} are popular to explain the general behaviors of image classifiers. It treats the input as trainable variables and optimizes the input via back-propagation to obtain abstract images to explain the model. However, the discrete adjacency matrices cannot be optimized in the same manner. In addition, several methods~\cite{dabkowski2017real, chen2018learning} learn soft masks to capture important image regions. However, applying soft masks to the adjacency matrices will destroy the discreteness property. \newcolor{Furthermore, graph nodes and edges contribute together to the final predictions of GNNs and the interactions between them are also important, which cannot be captured existing methods from image domain.}

In addition, for images and texts, we study the importance of each pixel or word. However, it is more important to study the structural information for graph data. First, the nodes in graphs may be unlabeled and the labels of the whole graphs are determined by graph structures. Then studying each node may be meaningless since those unlabeled nodes contain no semantic meaning. In addition, for graphs in biochemistry, neurobiology, ecology, and engineering, graph substructures are highly related to their functionalities~\cite{alon2007network}. An example is network motifs, which can be considered as the building blocks of many complex networks~\cite{milo2002network, alon2019introduction}. Then 
such structural information should not be ignored in the explanation tasks. However, existing methods from image domains cannot provide explanations regarding the
structures.  
Next, for node classification tasks, the prediction of each node is determined by different message walks from its neighbors. Investigating such message walks is meaningful but challenging. None of the existing methods in the image domain can consider such walk information, which needs further explorations.

Furthermore, graph data are less intuitive than images and texts. To understand deep models, domain knowledge for the datasets is necessary. For images and texts, understanding the semantic meaning of input data is simple and straightforward. For their explanations, humans can easily understand them even though the explanations are highly abstract. 
However, since graphs can represent complex data, such as molecules, social networks, and citation networks, it is challenging for humans to understand the meaning of graphs. In addition, in interdisciplinary areas such as chemistry and biology, there are many unsolved mysteries and the domain knowledge is still lacking. 
Hence, it is non-trivial to obtain human-understandable explanations for graph models, thus raising the need of standard datasets and evaluation metrics for explanation tasks. 

\begin{figure*}[ht!]
    \centering
    \includegraphics[width=2\columnwidth]{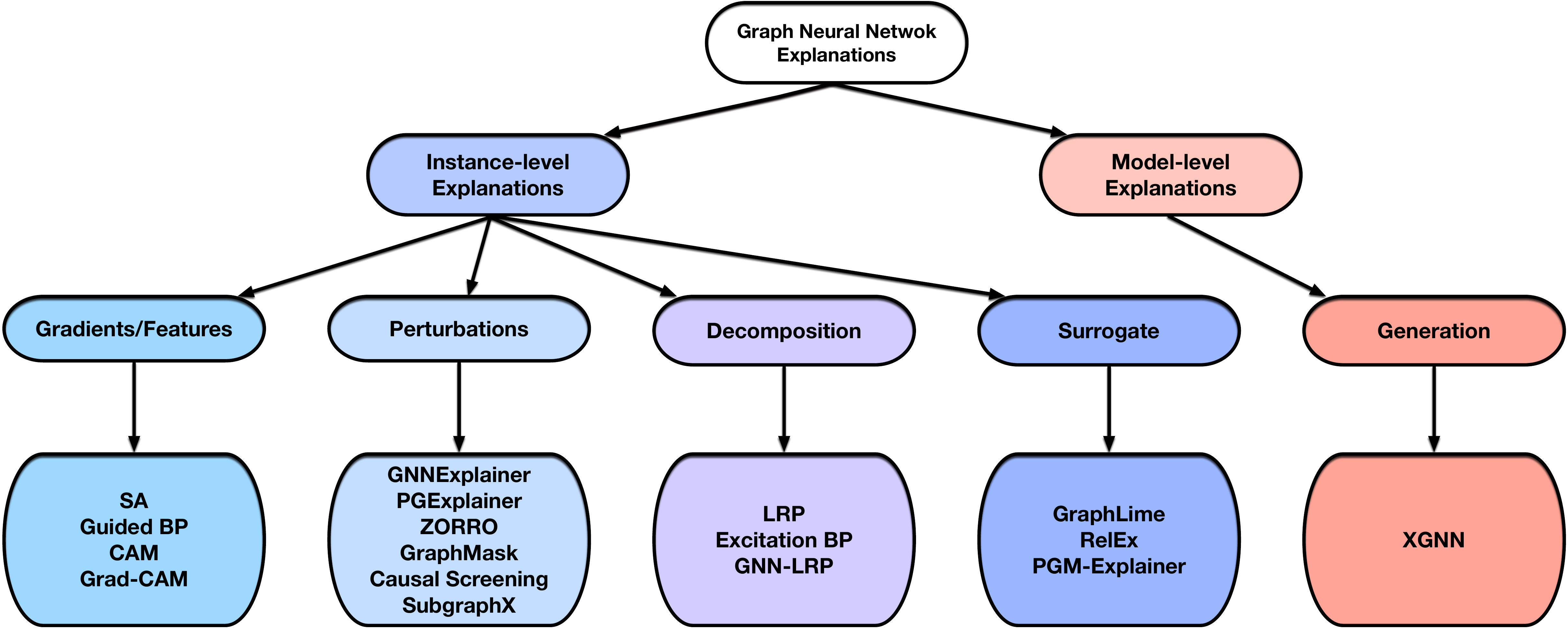}
    \caption{An overview of our proposed taxonomy. We categorize existing GNN explanation approaches into two branches: instance-level methods and model-level methods. For the instance-level methods, the gradients/features-based methods include SA~\cite{baldassarre2019explainability}, Guided BP~\cite{baldassarre2019explainability}, CAM~\cite{pope2019explainability}, and Grad-CAM~\cite{pope2019explainability}; the perturbation-based methods are GNNExplainer~\cite{ying2019gnnexplainer}, PGExplainer~\cite{luoparameterized}, ZORRO~\cite{anonymous2021hard}, GraphMask~\cite{schlichtkrull2020interpreting}, Causal Screening~\cite{anonymous2021causal}, and SubgraphX~\cite{yuan2021explainability}; the decomposition methods contains LRP~\cite{baldassarre2019explainability, schwarzenberg2019layerwise}, Excitation BP~\cite{pope2019explainability} and GNN-LRP~\cite{schnake2020higherorder}; the surrogate methods include GraphLime~\cite{huang2020graphlime}, RelEx~\cite{zhang2020relex}, and PGM-Explainer~\cite{vu2020pgm}. For the model-level methods, the only existing approach is XGNN~\cite{xgnn}.}
    \label{fig:overview}
\end{figure*}

\section{An Overview of the Taxonomy}
Recently, several approaches are proposed to explain the predictions of deep graph models. These methods focus on different aspects of the graph models and provide different views to understand these models. They generally answer a few questions; some of those are, which input edges are more important? which input nodes are more important? which node features are more important? what graph patterns will maximize the prediction of a certain class? To better understand these methods, we provide a taxonomy of different explanation techniques for GNNs. The structure of our taxonomy is shown in Figure~\ref{fig:overview}. Based on what types of explanations are provided, different techniques are categorized into two main classes: instance-level methods and model-level methods. 

First, instance-level methods provide input-dependent explanations for each input graph. Given an input graph, these methods explain deep models by identifying important input features for its prediction. Based on how the importance scores are obtained, we categorize the methods into four different branches: gradients/features-based methods, perturbation-based methods, decomposition methods, and surrogate methods. Specifically, the gradients/features-based methods employ the gradients or the feature values to indicate the importance of different input features. In addition, perturbation-based methods monitor the change of prediction with respect to  different input perturbations to study input importance scores. The decomposition methods first decompose the prediction scores, such as predicted probabilities, to the neurons in the last hidden layer. Then they back propagate such scores layer by layer until the input space and use such decomposed scores as the importance scores. Meanwhile, for a given input example, surrogate based methods first sample a dataset from the neighbors of the given example. Next, these methods fit a simple and interpretable model, such as the decision tree, to the sampled dataset. Then the explanations from the surrogate model are used to explain the original predictions. Second, model-level methods explain graph neural networks without respect to any specific input example. The input-independent explanations are high-level and explain general behaviors. Compared with instance-level methods, this direction is still less explored. The only existing model-level method is XGNN~\cite{xgnn}, which is based on graph generation. 

Overall, these two categories of methods explain deep graph models from different views. Instance-level methods provide example-specific explanations while model-level methods provide high-level insights and a generic understanding. To verify and trust deep graph models, it requires human supervision to check the explanations. For instance-level methods, more human supervisions are needed since experts need to explore the explanations for different input graphs. For model-level methods, since the explanations are high-level, less human supervisions are involved. Furthermore, the explanations of instance-level methods are based on real input instances so that they are easier to understand. However, the explanations for model-level methods may not be human-intelligible since 
the obtained graph patterns may not even exist in the real-world. Overall, these two types of methods can be combined together to better understand deep graph models, and hence it is necessary to investigate both of them.

\section{Instance-Level Explanations}
We start our explorations by introducing the instance-level methods for GNN explanations. In this section, we will discuss the motivations, methodologies, advantages, limitations of each category.

\subsection{Gradients/Features-Based Methods}
First, we introduce the gradients/features-based methods for explaining deep graph models.

\subsubsection{A unified view}
Employing gradients or features to explain the deep models is the most straightforward solution, which is widely used in image and text tasks. The key idea is to use the gradients or hidden feature map values as the approximations of input importance. Specifically, gradients-based methods~\cite{simonyan2013deep, smilkov2017smoothgrad} compute the gradients of target prediction with respect to input features by back-propagation. 
Meanwhile, features-based methods~\cite{zhou2016learning, selvaraju2017grad} map the hidden features to the input space via interpolation to measure importance scores. Generally, in such methods, larger gradients or feature values indicate higher importance. Note that both gradients and hidden features are highly related to the model parameters, then such explanations can reflect the information contained in the model. Since these methods are simple and general, they can be easily extended to the graph domain. Recently, 
several methods have been applied to explain graph neural networks, including  SA~\cite{baldassarre2019explainability}, Guided BP~\cite{baldassarre2019explainability}, CAM~\cite{pope2019explainability}, and Grad-CAM~\cite{pope2019explainability}.
The key difference among these methods lies in the procedure of gradient back-propagation and how different hidden feature maps are combined.

\subsubsection{Methods}
\textbf{SA}~\cite{baldassarre2019explainability} directly employs the squared values of gradients as the importance scores of different input features. It can be directly computed by back-propagation, which is the same as network training but the target is input features instead of model parameters. 
Note that the input features can be graph nodes, edges, or node features. It assumes that higher absolute gradient values indicate the corresponding input features are more important. While it is simple and efficient, it has several additional limitations. First, SA can only reflect the sensitivity between input and output, which cannot accurately show the importance. In addition, it also suffers from saturation problems \cite{shrikumar2017learning}. In the saturation regions of the model, where the model output changes minimally with respect to any input change, the gradients can hardly reflect the contributions of inputs. 
 
\textbf{Guided BP}~\cite{baldassarre2019explainability} shares a similar idea with SA but modifies the procedure of 
backs propagating gradients. Since negative gradients are challenging to explain, Guided BP only back propagates positive gradients while clipping negative gradients to zeros. Then only positive gradients are used to measure the importance of different input features. Note that Guided BP shares the same limitations as SA. 

\textbf{CAM}~\cite{pope2019explainability} maps the node features in the final layer to the input space to identify important nodes. It requires the GNN model to employ a global average pooling (GAP) layer and a fully-connected (FC) layer as the final classifier. 
Specifically, CAM takes the final node embeddings and combines different feature maps by weighted summations to obtain importance scores for input nodes. 
Note that the weights are obtained from the final fully-connected (FC) layer connected with the target prediction. This approach is also very simple but still has several major limitations. First, CAM has special requirements for the GNN structure, which limits its application and generalization. Second, it assumes that the final node embeddings can reflect the input importance, which is heuristic and may not be true. \newcolortwo{Furthermore, it can only explain graph classification models and cannot be directly applied to node classification tasks, since it needs the final FC layer to map predictions to different nodes.}

\textbf{Grad-CAM} \cite{pope2019explainability} extends the CAM to general graph classification models by removing the constraint of the GAP layer. Instead of using the weights between the GAP output and FC output, it employs gradients as the weights to combine different feature maps. Specifically, it first computes the gradients of the target prediction with respect to the final node embeddings. Then it averages such gradients to obtain the weight for each feature map. 
However, it is also based on heuristic assumptions and cannot explain node classification models.

\begin{figure*}[ht!]
    \centering
    \includegraphics[width=2\columnwidth]{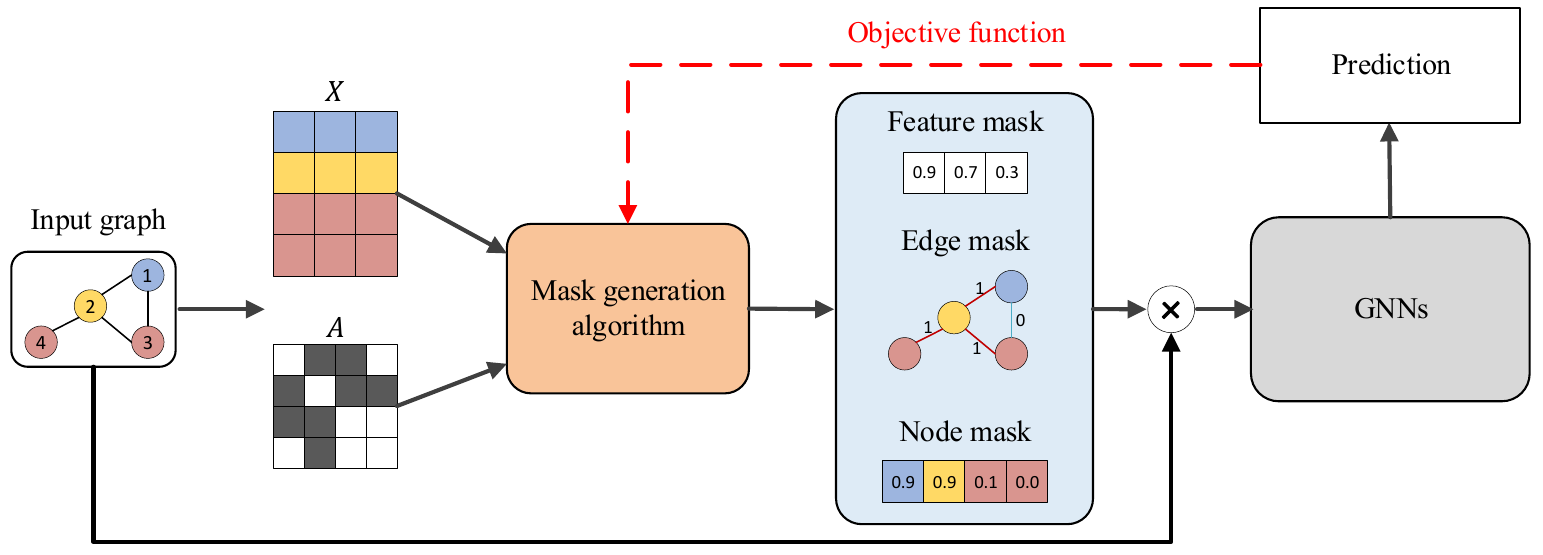}
    \caption{The general pipeline of the perturbation-based methods. They employ different mask generation algorithms to obtain different types of masks. Note that the mask can correspond to nodes, edges, or node features. In this example, we show a soft mask for node features, a discrete  mask for edges, and an approximated discrete mask for nodes.  Then the mask is combined with the input graph to capture important input information. Finally, the trained GNNs evaluate whether the new prediction is similar to the original prediction and can provide guidance for improving the mask generation algorithms. }
    \label{fig:perturbation}
\end{figure*}
\subsection{Perturbation-Based Methods}
Next, we introduce the perturbation-based methods for explaining deep graph models. 
\subsubsection{A unified view}
The perturbation-based methods are widely employed to explain deep image models~\cite{dabkowski2017real, pmlr-v80-chen18j, yuan2020interpreting}. The underlying motivation is to study the output variations with respect to different input perturbations. Intuitively, when important input information is retained, the predictions should be similar to the original predictions. Existing methods~\cite{dabkowski2017real, pmlr-v80-chen18j, yuan2020interpreting} learn a generator to generate a mask to select important input pixels to explain deep image models. However, such methods cannot be directly applied to graph models. Unlike images, graphs are represented as nodes and edges, and they cannot be resized to share the same node and edge numbers. In addition, different from images, the structural information is critical for graphs and can determine the functionalities.

To explain deep graph models, several perturbation-based methods are proposed, including GNNExplainer~\cite{ying2019gnnexplainer}, PGExplainer~\cite{luoparameterized}, ZORRO~\cite{anonymous2021hard}, GraphMask~\cite{schlichtkrull2020interpreting}, Causal Screening~\cite{anonymous2021causal}, and SubgraphX~\cite{yuan2021explainability}. They share a similar high-level pipeline as shown in Figure~\ref{fig:perturbation}. \newcolor{First, given the input graph, masks are obtained in different ways to indicate important input features.} Note that depending on the 
explanation tasks, different masks are generated, such as node masks, edge masks, and node feature masks. Next, the generated masks are combined with the input graph to obtain a new graph containing important input information. Finally, the new graph is fed into the trained GNNs to evaluate the masks and update the mask generation algorithms. Intuitively, the important input features captured by masks should convey the key semantic meaning and hence lead to a similar prediction to the original one.  
The difference between these methods mainly lies in three aspects: the mask generation algorithm, the type of masks, and the objective function.

It is noteworthy that there are three different types of masks: soft masks, discrete masks, and approximated discrete masks. In Figure~\ref{fig:perturbation}, we show soft masks for node features, discrete masks for edges, and approximated discrete masks for nodes.
The soft masks contain continuous values between $[0,1]$ and the mask generation algorithm can be directly updated by back-propagation. However, soft masks are suffered from the ``introduced evidence'' problem~\cite{dabkowski2017real} that any non-zero or non-one value in the mask may introduce new semantic meaning or new noise to the input graph, thus affecting the explanation results.  
Meanwhile, the discrete masks only contain discrete values $0$ and $1$, which can avoid the ``introduced evidence'' problem since no new numerical value is introduced. However, discrete masks always involve non-differentiable operations, such as sampling. One popular way to solve it is the policy gradient technique~\cite{sutton1999policy}. Furthermore, recent studies~\cite{chen2018learning, jang2016categorical, louizos2017learning} propose to employ reparameterization tricks, such as Gumbel-Softmax 
estimation and sparse relaxations, to approximate the discrete masks. Note that the output mask is not strictly discrete but provides a good approximation, which not only enables the back-propagation but also largely alleviates the ``introduced evidence'' problem.

\subsubsection{Methods}
\textbf{GNNExplainer}~\cite{ying2019gnnexplainer} learns soft masks for edges and node features to explain the predictions via mask optimization. \newcolor{To obtain masks, it randomly initializes soft masks and treats them as trainable variables.} Then GNNExplainer combines the masks with the original graph via element-wise multiplications. Next, the masks are optimized by maximizing the mutual information between the predictions of the original graph and the predictions of the newly obtained graph. Even though different regularization terms, such as element-wise entropy, are employed to encourage optimized masks to be discrete, the obtained masks are still soft masks so that GNNExplainer cannot avoid the ``introduced evidence'' problem. In addition, the masks are optimized for each input graph individually and hence the explanations may lack a global view. 

\textbf{PGExplainer}~\cite{luoparameterized} learns approximated discrete masks for edges to explain the predictions. \newcolor{To obtain edge masks, it trains a parameterized mask predictor to predict edge masks.} Given an input graph, it first obtains the embeddings for each edge by concatenating the corresponding node embeddings. Then the predictor uses the edge embeddings to predict the probability of each edge being selected, which can be treated as the importance score. Next, the approximated discrete masks are sampled via the reparameterization trick. Finally, the mask predictor is trained by maximizing the mutual information between the original predictions and new predictions. Note that even though the reparameterization trick is employed, the obtained masks are not strictly discrete but can largely alleviate the ``introduced evidence'' problem. In addition, since all edges in the dataset share the same predictor, the explanations can provide a global understanding of the trained GNNs.

\textbf{GraphMask}~\cite{schlichtkrull2020interpreting} is a post-hoc method for explaining the edge importance in each GNN layer. Similar to the PGExplainer, it trains a classifier to predict whether an edge can be dropped without affecting the original predictions. However, GraphMask obtains an edge mask for each GNN layer while PGExplainer only focuses the input space. In addition, to avoiding changing graph structures, the dropped edges are replaced by learnable baseline connections, which are vectors with the same dimensions as node embeddings. Note that binary Concrete distribution~\cite{louizos2017learning} and reparameterization trick are employed to approximate discrete masks. In addition, the classifier is trained using the whole dataset by minimizing a divergence term, which measures the difference between network predictions. Similar to PGExplainer, it can largely alleviate the ``introduced evidence'' problem and provide a global understanding of the trained GNNs.

\begin{figure*}[ht!]
    \centering
    \includegraphics[width=2\columnwidth]{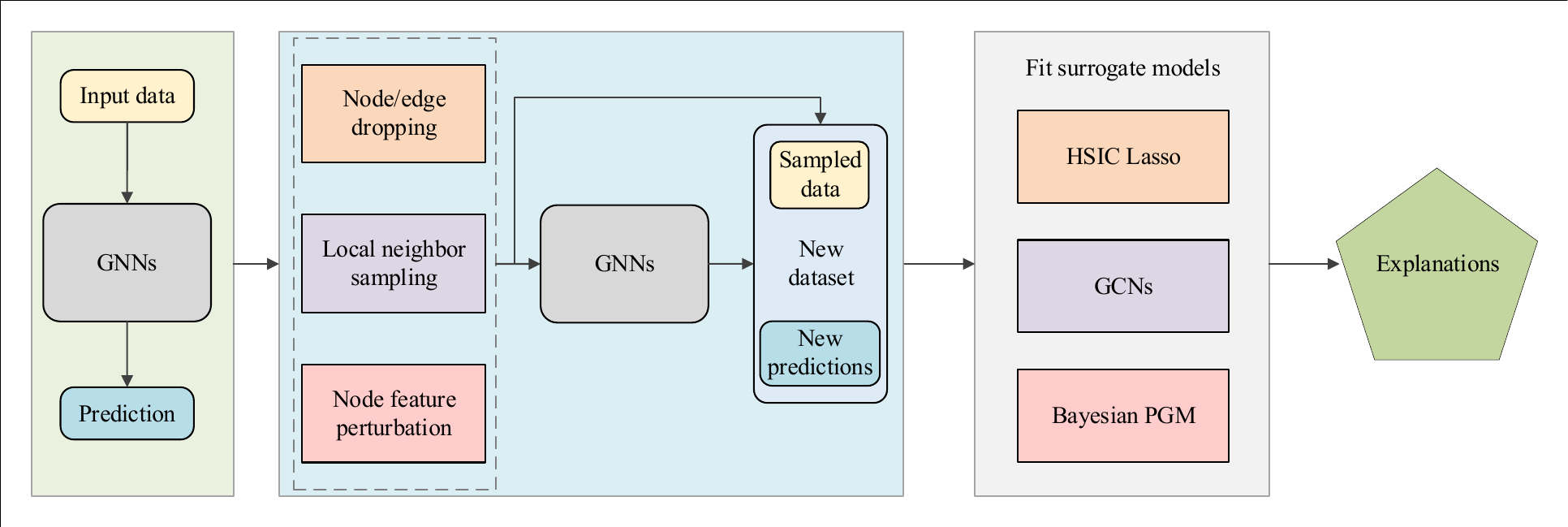}
    \caption{The general pipeline of the surrogate methods. Given an input graph and its prediction, they first sample a local dataset to represent the relationships around the target data. Then different surrogate methods are applied to fit the local dataset. Note that surrogate models are generally simple and interpretable ML models. Finally, the explanations from the surrogate model can be regarded as the explanations of the original prediction. }
    \label{fig:surrogate}
\end{figure*}
\textbf{ZORRO}~\cite{anonymous2021hard} employs discrete masks to identify important input nodes and node features. \newcolor{Given an input graph, a greedy algorithm is used to select nodes or node features step by step to obtain discrete masks for nodes and features.} For each step, ZORRO selects one node or one node feature with the highest fidelity score. Note that the objective function, fidelity score, measure how the new predictions match the original predictions of the model by fixing the selected nodes/features and replacing the others with random noise values. Since there is no training procedure involved, the non-differentiable limitation of discrete masks is avoided. In addition, by using hard masks, ZORRO is not suffered from the ``introduced evidence'' problem. However, the greedy mask selection algorithm may lead to local optimal explanations. In addition, the explanations may lack a global understanding since masks are generated for each graph individually.

\textbf{Causal Screening}~\cite{anonymous2021causal} studies the causal attribution of different edges in the input graph as explanations. It identifies an edge mask for the explanatory subgraph. The key idea of causal attribution is to study the change of predictions when adding an edge into the current explanatory subgraph, known as the causal effect. \newcolor{To obtain edge masks for each step, it studies the causal effects of different edges and selects the edge with the highest causal effect to add to the explanatory subgraph.} Specifically, it employs the individual causal effect (ICE) to select edges, which measures mutual information (between the predictions of original graphs and the explanatory subgraphs)  difference after adding different edges to the subgraph. Similar to ZORRO, Causal Screening is a greedy algorithm generating discrete masks without any training procedure. Hence, it is not suffered from the ``introduced evidence'' problem but may lack a global understanding and stuck in local optimal explanations. 

\textbf{SubgraphX}~\cite{yuan2021explainability} explores subgraph-level explanations for deep graph models. It employs the Monte Carlo Tree Search (MCTS)  algorithm~\cite{silver2017mastering} to efficiently
explore different subgraphs via node pruning and select the most important subgraph from the leaves of the search tree as the explanation of the prediction. In addition, it employs Shapley value~\cite{kuhn1953contributions} as the reward of MCTS to measure the importance of subgraphs and proposes an efficient approximation of Shapley value by only considering the interactions within message passing range. While SubgraphX is not directly studying masks,
the MCTS algorithm can be understood as the mask generation algorithm and its node pruning actions can be regarded as different masks to obtain subgraphs. In addition, the Shapley values can be treated as the objective function to update the mask generation algorithm. Compared with other perturbation-based methods, its obtained subgraphs are more human-intelligible and suitable for graph data.  However, the computational cost is more expensive since it needs to explore different subgraphs with the MCTS algorithm.

\subsection{Surrogate Methods}
In this section, we introduce the surrogate methods for explaining deep graph models. 
\subsubsection{A unified view}
Deep models are challenging to explain because of the complex and non-linear relationships between the input space and output predictions. A popular way to provide instance-level explanations for image models is known as surrogate method. The underlying idea is to employ a simple and interpretable surrogate model to approximate the predictions of the complex deep model for the neighboring areas of the input example. Note that these methods assume that the relationships in the neighboring areas of the input example is less complex and can be well captured by a simpler surrogate model.  
Then the explanations from the interpretable surrogate model are used to explain the original prediction. Applying surrogate methods to the graph domain is challenging since graph data are discrete and contain topology information. Then it is not clear how to define the neighboring regions of the input graph and what interpretable surrogate models are suitable.

Recently, several surrogate methods are proposed to explain deep graph models, including GraphLime~\cite{huang2020graphlime}, RelEx~\cite{zhang2020relex}, and PGM-Explainer~\cite{vu2020pgm}. The general pipeline of these methods is shown in Figure~\ref{fig:surrogate}.
To explain the prediction of a given input graph, they first obtain
a local dataset containing multiple neighboring data objects and their predictions. Then they fit a interpretable model to learn the local dataset. Finally, the explanations from the interpretable model are regarded as the explanations of the original model for the input graph. While these methods share a similar high-level idea, the key difference lies in two aspects: how to obtain the local dataset and what interpretable surrogate model to use. 

\subsubsection{Methods}
\textbf{GraphLime}~\cite{huang2020graphlime} extends the LIME algorithm~\cite{todoriki2021semi, ribeiro2016should} to deep graph models and studies 
the importance of different node features for node classification tasks. 
Given a target node in the input graph, GraphLime considers its $N$-hop neighboring nodes and their predictions as its local dataset where a reasonable choice of $N$ is the number of layers in the trained GNNs. Then a nonlinear surrogate model, Hilbert-Schmidt Independence Criterion (HSIC) Lasso~\cite{yamada2014high}, is employed to fit the local dataset. Note that HSIC Lasso is a kernel-based feature selection algorithm. Finally, based on the weights of different features in  HSIC Lasso, it can select important features to explain the HSIC Lasso predictions. Those selected features are regarded as the explanations of the original GNN prediction. 
However, GraphLime can only provide explanations for node features but ignore graph structures, such as nodes and edges, which are more important for graph data. In addition, GraphLime is proposed to explain node classification predictions but cannot be directly applied to graph classification models. 

\begin{figure*}[ht!]
    \centering
    \includegraphics[width=2\columnwidth]{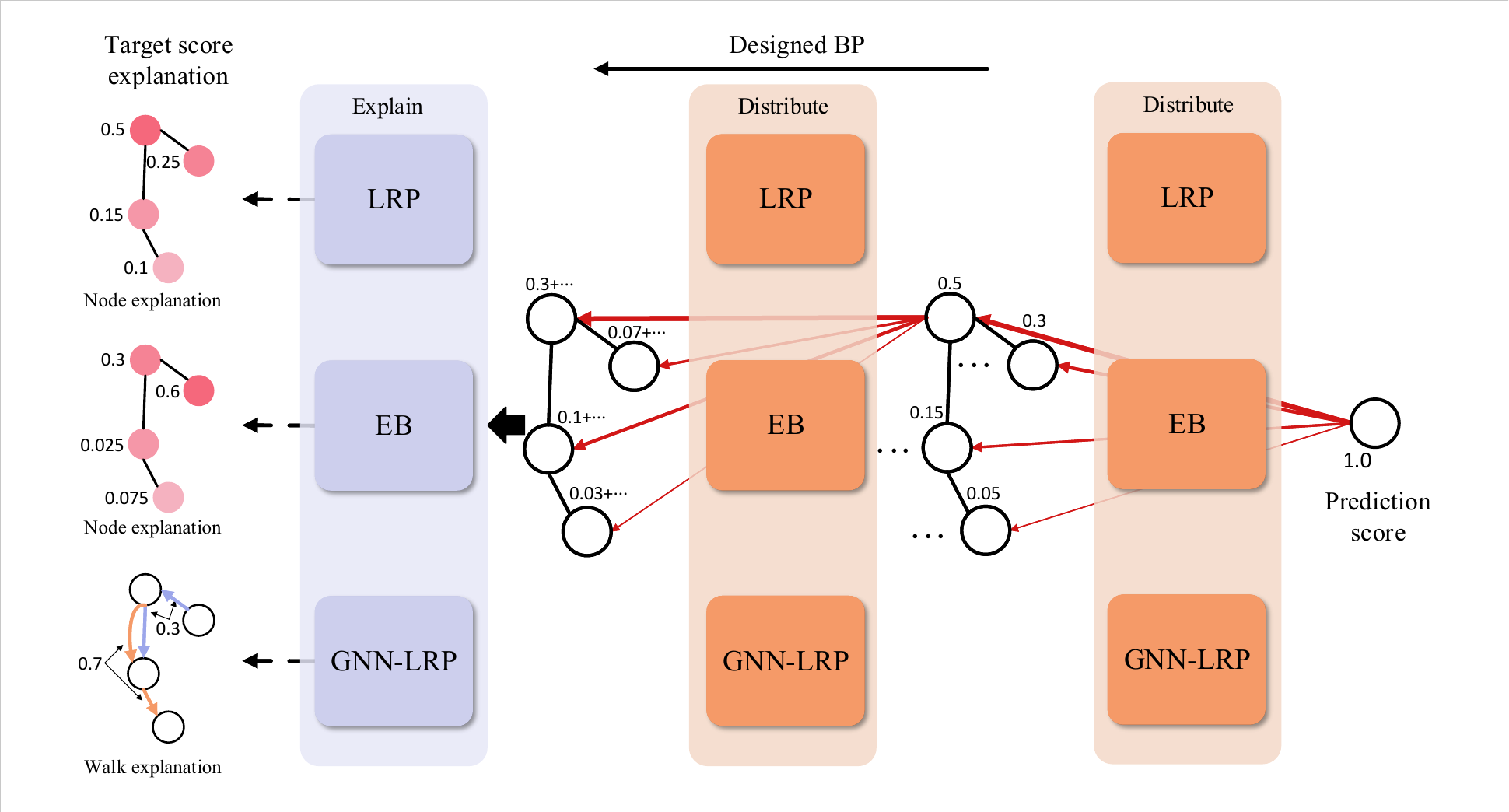}
    \caption{The general pipeline of the decomposition methods. These methods distribute the prediction score to input space to indicate the input importance. The target score is decomposed layer by layer in a back-propagation manner. Note that the main difference among these methods is the employed score decomposition rule.}
    \label{fig:decomposition}
\end{figure*}
\textbf{RelEx}~\cite{zhang2020relex} also studies the explainability of node classification models by combining the ideas of surrogate methods and perturbation-based methods. Given a target node and its computational graph ($N$-hop neighbors), it first obtains a local dataset by randomly sampling connected subgraphs from the computational graph and feeding these subgraphs to the trained GNNs to obtain their predictions. 
Specifically, starting from the target node, it randomly selects neighboring nodes in a BFS manner. Next, it employs a GCN model as the surrogate model to fit the local datasets. Note that different from LIME and GraphLime, the surrogate model in RelEx is not interpretable. After training, it further applies the aforementioned perturbation-based methods, such as generating soft masks or Gumbel-Softmax masks, to explain the predictions. Compared with GraphLime, it can provide explanations regarding important nodes. However, it contains multiple steps of approximations, such as using the surrogate model to approximate local relationships and using masks to approximate the edge importance, thus making the explanations less convincing and trustable. Furthermore, as perturbation-based methods can be directly employed to explain original deep graph models, then it is not necessary to build another non-interpretable 
deep model as the surrogate model to explain. It is also unknown how it can be applied for graph classification tasks.

\textbf{PGM-Explainer}~\cite{vu2020pgm} builds a probabilistic graphical model to provide instance-level explanations for GNNs. The local dataset is obtained by random node feature perturbation. Specifically, given an input graph, each time PGM-Explainer randomly perturbs the node features of several random nodes within the computational graph. Then for any node in the computational graph, PGM-Explainer records a random variable indicating whether its features are perturbed and its influence on the GNN predictions. By repeating such procedures multiple times, a local dataset is obtained. Note that the local dataset of PGM-Explainer contains node variables instead of different neighboring graph samples. Then it selects top dependent variables to reduce the size of the local dataset via the Grow-Shrink (GS) algorithm~\cite{margaritis1999bayesian}. Finally, an interpretable Bayesian network is employed to fit the local dataset and to explain the predictions of the original GNN model. PGM-Explainer can provide explanations regarding graph nodes but ignore graph edges, which contain important graph topology information. In addition, different from GraphLime and RelEx, the PGM-Explainer can be used to explain both node classification and graph classification tasks.

\subsection{Decomposition Methods}

Finally, we introduce the decomposition methods for explaining deep graph models.

\subsubsection{A unified view}

Another popular way to explain deep image classifiers is decomposition methods, which measure the importance of input features by decomposing the original model predictions into several terms. Then those terms are regarded as the importance scores of the corresponding input features. 
These methods directly study the model parameters to reveal the relationships between the features in the input space and the output predictions.  Note that
the conservative property of these methods requires that the sum of the decomposed terms is equal to the original prediction score. However, it is challenging to directly apply such methods to the graph domain since graphs contain nodes, edges, and node features. It is non-trivial to distribute scores to different edges while graph edges contain important structural information that cannot be ignored.

Recently, several decomposition methods are proposed to explain the deep graph neural networks, including Layer-wise Relevance Propagation (LRP)~\cite{baldassarre2019explainability, schwarzenberg2019layerwise}, Excitation BP~\cite{pope2019explainability} and GNN-LRP~\cite{schnake2020higherorder}. The intuition of these algorithms is to build score decomposition rules to distribute the prediction scores to the input space. The general pipeline of these methods is shown in Figure~\ref{fig:decomposition}. These methods distribute the prediction score layer by layer in a back-propagation manner until the input layer. 
Starting from the output layer, the model's prediction is treated as the initial target score. Then the score is decomposed and distributed to the neurons in the previous layer following the decomposition rules. By repeating such procedures until input space, they can obtain importance scores for node features, which can be combined to represent edge importance, node importance, and walk importance. 
It is noteworthy that all of these algorithms ignore the activation functions in deep graph models. \newcolor{The main difference among these methods is the score decomposition rules, the view of decomposition, and the targets of explanations.}



\subsubsection{Methods}

\textbf{LRP}~\cite{baldassarre2019explainability, schwarzenberg2019layerwise} extends the original LRP algorithm~\cite{bach2015pixel} to deep graph models.
It decomposes the output prediction score to different node importance scores. The score decomposition rule is developed based on the hidden features and weights. For a target neuron, its score is represented as a linear approximation of neuron scores from the previous layer. Intuitively, the neuron with a higher contribution of the target neuron activation receives a larger fraction of the target neuron score. 
Specifically, \etal{Baldassarre}~\cite{baldassarre2019explainability} applies the $\epsilon$-stabilized rule from the original LRP, while \etal{Schwarzenberg}~\cite{schwarzenberg2019layerwise} employs the $z^+$-rule to perform decomposition. 
To satisfy the conservative property, the adjacency matrix is treated as a part of GNN model in the post-hoc explanation phase so that it can be ignored during score distribution; otherwise, the adjacent matrix will also receive decomposed scores, thus making the conservative property invalid. Since LRP is directly developed based on the model parameters, its explanation results are more trustable. However, it can only study the importance of different nodes and cannot be applied to graph structures, such as subgraphs and graph walks, which are more important for understanding GNNs. In addition, such a algorithm requires a comprehensive understanding of the model structures, which limits its applications for non-expert users, such as interdisciplinary researchers.

\newcolor{\textbf{Excitation BP}~\cite{pope2019explainability} shares a similar idea as the LRP algorithm but is developed from the view of the law of total probability.} It 
defines that the probability of a neuron in the current layer is equal to the total probabilities it outputs to all connected neurons in the next layer. Then the score decomposition rule can be regarded as decomposing the target probability into several conditional probability terms. Note that the computation of Excitation BP is highly similar to the $z^+$-rule in LRP. Hence, it shares the same advantages and limitations as the LRP algorithm.


\textbf{GNN-LRP}~\cite{schnake2020higherorder} studies the importance of different graph walks. It is more coherent to the deep graph neural networks since graph walks correspond to message flows when performing neighborhood information aggregation. \newcolor{It treats the GNN model as a function and provides a view of high-order Taylor decomposition to develop the score decomposition rule.
 It is theoretically proven that the Taylor decomposition (at root zero) only contains $T$-order terms where $T$ is the number of layers in the trained GNNs and each term corresponds to a $T$-step graph walk. Then such terms can be regarded as the importance scores of graph walks.} Since it is not possible to directly compute the high-order derivatives given by Taylor expansion, GNN-LRP also follows a back propagation procedure to approximate the $T$-order terms. 
Note that the back propagation computation in GNN-LRP is similar to the LRP algorithm. However, instead of distributing scores to nodes or edges, GNN-LRP distributes scores to different graph walks. 
It records the paths of the distribution processes from layer to layer. Those paths are considered as different walks and the scores are obtained from their corresponding nodes. 
While GNN-LRP has a solid theoretical background, the approximations in its computations may not be accurate. In addition, the computational complexity is high since each walk is considered separately. \newcolor{Furthermore, it is also challenging for non-experts to use, especially for interdisciplinary domains, because the back propagation rules need to be derived for different models and GNN variants.}




\begin{table*}
    \caption{A  comprehensive analysis of different explanation methods. Here ``Type'' indicates what type of explanations are provided, ``Learning'' denotes whether learning procedures are involved, ``Task'' means what tasks each method can be applied to, ``Target'' indicates the targets of explanations, ``Black-box'' means if the trained GNNs are treated as a black-box during the explanation stage, ``Flow'' denotes the computational flow for explanations, and ``Design'' indicates whether an explanation method has specific designs for graph data. Note that GC denotes graph classification, NC denotes node classification, N means nodes, E means edges, NF 
   represents node features, and Walk indicates graph walks. }
    \label{Analysis}
    \centering
    \begin{tabular}{@{}lccccccc@{}}
        \toprule[1.5pt]
        \textbf{Method}  & \textsc{Type}     & \textsc{Learning}& {\textsc{Task}} & {\textsc{Target}} &  {\textsc{Black-box}} & {\textsc{Flow}} & {\textsc{Design}}  \\
        \midrule[1pt]
        \textbf{SA~\cite{baldassarre2019explainability, pope2019explainability}}  & Instance-level  & \xmark & GC/NC &N/E/NF & \xmark & Backward  &\xmark   \\
        \textbf{Guided BP~\cite{baldassarre2019explainability}}      & Instance-level & \xmark  &GC/NC& N/E/NF& \xmark& Backward &\xmark    \\
        \textbf{CAM~\cite{pope2019explainability} }      & Instance-level & \xmark  & GC & N & \xmark& Backward   &\xmark   \\
        \textbf{Grad-CAM~\cite{pope2019explainability}} & Instance-level & \xmark  & GC & N & \xmark& Backward  &\xmark \\
        \midrule
        \textbf{GNNExplainer~\cite{ying2019gnnexplainer}} & Instance-level & \cmark & GC/NC & E/NF & \cmark& Forward  &\cmark \\
        \textbf{PGExplainer~\cite{luoparameterized}} & Instance-level & \cmark  &GC/NC & E & \xmark& Forward &\cmark  \\
        \textbf{GraphMask~\cite{schlichtkrull2020interpreting}} & Instance-level & \cmark  &GC/NC & E & \xmark& Forward &\cmark  \\
        \textbf{ZORRO~\cite{anonymous2021hard}} & Instance-level & \xmark  &GC/NC & N/NF & \cmark& Forward &\cmark  \\
        \textbf{Causal Screening~\cite{anonymous2021causal}} & Instance-level & \xmark  &GC/NC & E & \cmark& Forward &\cmark \\
        \textbf{SubgraphX~\cite{yuan2021explainability}} & Instance-level & \cmark  &GC/NC & Subgraph & \cmark& Forward &\cmark \\
        \midrule
        \textbf{LRP~\cite{baldassarre2019explainability, schwarzenberg2019layerwise}} & Instance-level & \xmark &GC/NC & N & \xmark& Backward  &\xmark \\
        \textbf{Excitation BP~\cite{pope2019explainability}} & Instance-level & \xmark & GC/NC & N & \xmark& Backward &\xmark \\
        \textbf{GNN-LRP~\cite{schnake2020higherorder}} & Instance-level & \xmark &GC/NC & Walk & \xmark& Backward  &\cmark \\
        \midrule
        \textbf{GraphLime~\cite{huang2020graphlime}} & Instance-level & \cmark & NC & NF & \cmark& Forward &\xmark  \\
        \textbf{RelEx~\cite{zhang2020relex}} & Instance-level & \cmark &NC & N/E & \cmark& Forward &\cmark  \\
        \textbf{PGM-Explainer~\cite{vu2020pgm}} & Instance-level & \cmark &GC/NC & N & \cmark& Forward &\cmark  \\
        \midrule
        \textbf{XGNN~\cite{xgnn}} & Model-level & \cmark & GC & Subgraph & \cmark& Forward&\cmark   \\
        \bottomrule[1pt]
    \end{tabular}
\end{table*}

\section{Model-level Explanations}
Different from instance-level methods, model-level methods aim at providing the general insights and high-level understanding to explain deep graph models. Specifically, they study what input graph patterns can lead to a certain GNN behavior, such as maximizing a target prediction.  
Input optimization~\cite{olah2017feature} is a popular direction to obtain model-level explanations for image classifiers. However, it cannot be directly applied to graph models due to the discrete graph topology information, thus making explaining GNNs at the model-level more challenging. Hence, it is still an important but less studied topic. To the best of our knowledge, the only existing model-level method for explaining graph neural networks is XGNN~\cite{xgnn}.

\textbf{XGNN} proposes to explain GNNs via graph generation. Instead of directly optimizing the input graph, it trains a graph generator so that the generated graphs can maximize a target graph prediction. Then the generated graphs are regarded as the explanations for the target prediction and are expected to contain discriminative graph patterns. In XGNN, the graph generation is formulated as a reinforcement learning problem. For each step, the generator predicts how to add an edge to the current graph. Then the generated graphs are fed into the trained GNNs to obtain feedback to train the generator via policy gradient. In addition, several graph rules are incorporated to encourage the explanations to be both valid and human-intelligible. \newcolor{For example, the rule for chemical data can be defined as the node degree of an atom should
not exceed its maximum chemical valency. In addition, XGNN can set a upper bond on the size of generated graphs so that the final explanations are succinct.} Note that XGNN is a general framework for generating model-level explanations so that any suitable graph generation algorithm can be applied. In addition, the explanations are general and provide a global understanding of the trained GNNs. However, the proposed XGNN only demonstrates its effectiveness in explaining graph classification models and it is unknown 
whether XGNN can be applied to node classification tasks, which is an important direction to explore in future studies.

\section{A Comprehensive Comparative Analysis}
In this section, we provide a comprehensive analysis of different explanation methods. The properties of different models are concluded in Table~\ref{Analysis}. Specifically, for each technique, we focus on the following aspects. 
\begin{itemize}
\item\textbf{Type:} It indicates the type of explanations provided by each explanation technique, including instance-level explanations and model-level explanations. Instance-level explanations are input-dependent and more specific while model-level explanations are more high-level and general. 
\item\textbf{Learning}: It denotes whether the explanation technique involves any learning procedure. Generally, the explanation method with learning procedures tends to better capture the relationships between inputs and predictions. However, they become less trustable
when the learning procedures introduce additional black-boxes. As shown in Table~\ref{Analysis}, there is no learning procedure in the feature/gradient-based methods and decomposition methods. 

\item\textbf{Task}: It indicates what tasks each method can be applied to. Here we only consider node classification (NC) and graph classification (GC) tasks. It demonstrates the generalizability of different methods. 

\item\textbf{Target}: It shows the target explanation of each method, such as showing node importance, edge importance, or graph walk importance. In Table~\ref{Analysis}, N denotes nodes, E means edges, NF denotes node features, and Walk denotes graph walks. For graph models, it is more important to study the structures of input graphs, such as edges, graph walks, and subgraphs. 

\item\textbf{Black-box}: It indicates whether the explanation technique treats the trained GNNs as a black-box. Several methods, including XGNN, GraphLime, PGM-Explainer, GNNExplainer, ZORRO, Causal Screening, and RelEx, only need to access the inputs and outputs to explain the trained GNNs, which means the trained GNNs are still treated as a black-box. For the other methods, they need to access the parameters or hidden representations of the trained GNNs. Note that by treating the trained GNNs as a black-box, the explanation method can be better generalized to applications where the GNN model is encapsulated or highly complex.  

\item\textbf{Flow}: It denotes the computational flow for explaining the trained GNNs. Several methods generate explanations in a backward manner, such as SA, Guided BP, CAM, Grad-CAM, LRP, Excitation BP, and GNN-LRP. The other methods only need the forward computations of the trained GNNs.

\item\textbf{Design}: It shows whether an explanation technique has specific designs for graph data or is simply extended from the image domain. Since graph data are special and their topology/structural information is important, it is necessary to explicitly consider such information to provide explanations.  \newcolor{For the methods that do not have specific designs for graphs, they cannot consider the topology information and structures of graph data, which may limit the quality of their explanations.}

\end{itemize}

\section{Evaluations}
In this section, we introduce the evaluations of graph model explanation techniques. Intuitively, good explanation results should faithfully explain the behaviors of GNN models. However, evaluating the explanation results is non-trivial due to the lack of ground truths. Hence, we discuss and analyze several commonly used datasets and metrics.  

\subsection{Datasets}
First, it is critical to select proper datasets to evaluate different explanation techniques. Since human evaluations are necessary, the explanations need to be understandable by humans. Hence, it is desired that the data are intuitive and easy to visualize. In addition, good datasets should contain human-understandable rationales between the data examples and labels so that experts can verify whether such rationales are identified by explanation algorithms. 
To evaluate different explanation techniques, several types of datasets are commonly employed, including synthetic data, sentiment graph data, and molecular data.

\subsubsection{Synthetic data}
Recently, several synthetic datasets are built to evaluate explanation techniques~\cite{ying2019gnnexplainer,luoparameterized}. In such datasets, different graph motifs are included and can determine the node labels or graph labels. In addition, the relationships between data examples and data labels are well defined by humans. Even though the trained GNNs may not perfectly capture such relationships, the graph motifs can be employed as reasonable approximations of the ground truths of explanation results. Here we introduce several common synthetic datasets. 

\begin{table}[ht!]

\caption{{Statistics and properties of our sentiment graph datasets. We report the prediction accuracies of three GNN models on our datasets. Note that ``BERT accuracy'' denotes the fine-tuning results of a pre-trained twelve-layer base BERT using original sentence data.}}

\centering
\begin{tabular}{l|ccc}
\hline 
Dataset &  Graph-SST2 & Graph-SST5 & Graph-Twitter \\
\hline 
\# of classes     & 2       & 5           & 3         \\
\# of features    & 768     & 768         & 768       \\
Avg. \# of nodes   & 10.199  & 19.849      & 21.103    \\
\# of train graphs     & 67,349  & 8,544  & 4,998  \\
\# of val. graphs   & 872     & 1,101  & 1,250  \\
\# of test graphs         & 1,821   & 2,210  & 692    \\\hline
GCN accuracy & 0.892 & 0.443 & 0.614 \\
GAT accuracy & 0.901 & 0.462 & 0.601 \\ 
GIN accuracy & 0.873 & 0.463 & 0.624 \\
\hline
BERT accuracy   & 0.912   & 0.532       & 0.736     \\ \hline
\end{tabular}
\label{table:Sentiment graph dataset}
\end{table}    
\begin{enumerate}
    \item BA-shapes: It is a node classification dataset with 4 different node labels. For each graph, it contains a base graph (300 nodes) and a house-like five-node motif. Note that the base graph is obtained by the \textit{Barabási-Albert} (BA) model, which can generate random scale-free networks with a preferential attachment mechanism~\cite{albert2002statistical}. The motif is attached to the base graph while random edges are added. Each node is labeled based on whether it belongs to the base graph or different spatial locations of the motif.  

    \item BA-Community: It is a node classification dataset with 8 different labels. For each graph, it is obtained by combining two BA-shapes graphs with randomly added edges. Node labels are determined by the memberships of BA-shapes graphs and their structural locations.

    \item Tree-Cycle: It is a node classification dataset with 2 different labels. For each graph, it consists of a base balanced tree graph with the depth equal to 8 and a six-node cycle motif. These two components are randomly connected. The label for the nodes in based graphs is 0 otherwise 1.

    \item Tree-Grids: It is a node classification dataset with 2 different labels. It is the same as the Tree-Cycle dataset, except that the Tree-Grids dataset employs nine-node grid motifs instead of cycle motifs.

    \item BA-2Motifs: It is a graph classification dataset with 2 different graph labels. There are 800 graphs and each of them is obtained by attaching different motifs, such as the house-like motif and the five-node cycle motif, to the base BA graph. Different graphs are labeled based on the type of motif. 
\end{enumerate}
 
Note that in these datasets, all node features are initialized as vectors containing all 1s. The trained GNNs models are expected to capture graph structures to make predictions. Then based on the rules of building each dataset, we can analyze the explanation results. For example, in the BA-2Motifs dataset, we can study whether the explanations can capture the motif structures. However, synthetic datasets only contain simple relationships between graphs and labels, which may be not enough for comprehensive evaluations.

\begin{figure*}[ht!]
    \centering
\includegraphics[width=2\columnwidth]{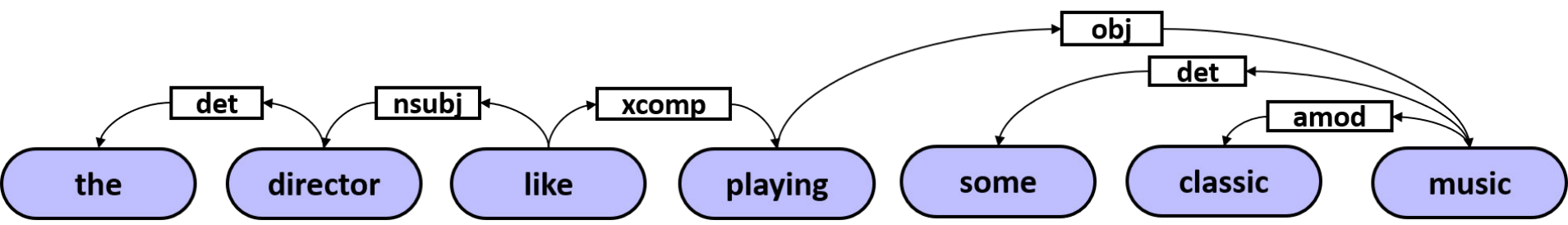}
    \caption{A data example of our graph sentiment datasets. Each node represents a word while edges indicates the relationship between the words. Note that edges are directed but their labels are ignored. }
    \label{fig:graph sentence structure}
\end{figure*}

\subsubsection{Sentiment graph data}
Traditional graph datasets are challenging to understand since humans only have limited domain knowledge. It raises the need of building human-understandable graph datasets. Text data can be a proper choice for graph explanation tasks since they consist of words and phrases with understandable semantic meanings. Then the explanation results can be easily evaluated by humans. 
Hence, we build three sentiment graph datasets based on text sentiment analysis data, including SST2~\cite{socher2013recursive}, SST5~\cite{socher2013recursive}, and  Twitter~\cite{dong2014adaptive} datasets. 

First, for each text sequence, we convert it to a graph that each node represents a word while the edges reflect the relationships between different words. Specifically, we employ the Biaffine parser~\cite{gardner2018allennlp} to extract word dependencies. We show an example of our generated sentiment graph in Figure~\ref{fig:graph sentence structure}. Note that the generated graphs are directed but the edge labels are ignored since most GNNs cannot capture edge label information. Next, we employ BERT~\cite{devlin2018bert} to obtain the word embeddings and such embeddings are used as initial embeddings of graph nodes. Specifically, we use a pre-trained twelve-layer base BERT model to extract a 768-dimension feature vector for each word. 


We build three sentiment graph datasets, known as Graph-SST2, Graph-SST5, and Graph-Twitter. They are now publicly available and can be directly used to study different explanation techniques\footnote{\url{https://github.com/divelab/DIG/tree/main/dig/xgraph/datasets}}. The statistics and properties of these datasets are reported in Table~\ref{table:Sentiment graph dataset}. We report the prediction accuracies of two-layer GNNs on these datasets, including GCNs, GATs, and GINs. In addition, we also show the 
fine-tuned accuracies of a pre-trained twelve-layer base BERT\cite{devlin2018bert} using the original sentence datasets. Since the results of BERT are obtained by finetuning the model and word embeddings, it is expected to perform better than the GNN models without finetuned node embeddings. Notably, we choose to use the word  embeddings as the node embeddings without finetuning because finetuning may encode the label information into the word embeddings, which  affect the structural information captured by GNNs and the final explanations. Hence, we believe these datasets are reasonable choices for building graph models to explain. 
Based on the semantic meaning of different words and the sentiment labels, we can study whether the explanations can identify the words with key meanings and the relationships among different words. 


\subsubsection{Molecule data}
Molecular datasets are also widely used in explanation tasks, such as MUTAG~\cite{debnath1991structure},  BBBP~\cite{martins2012bayesian}, and Tox21~\cite{wu2018moleculenet}. Each graph in such datasets corresponds to a molecule where nodes represent atoms and edges are the chemical bonds. The labels of molecular graphs are generally determined by the chemical functionalities or properties of the molecules. Employing such datasets for explanation tasks requires domain knowledge, such as what chemical groups are discriminative for their functionalities. For example, in the dataset MUTAG, different graphs are labeled based on their mutagenic effects on a bacterium. In addition, it is known that carbon rings and $NO_2$ chemical groups may lead to mutagenic effects~\cite{debnath1991structure}. Then we can study whether the explanations can identify such patterns for the corresponding class.

\subsection{Evaluation Metrics}
Even though visualization results can provide an insightful understanding regarding whether the explanations are reasonable to humans, such evaluations are not fully trustable due to the lack of ground truths. In addition, to compare different explanation methods, humans need to study the results for each input example, which is time-consuming. Furthermore, human evaluations are highly dependent on their subjective understanding, which is not fair enough. Hence, evaluation metrics are crucial for studying explanation
methods. Good metrics should evaluate the results from the model's perspective, such as whether the explanations are faithful to the model~\cite{faithfulness, wiegreffe2019attention}. In this section, we introduce several recently proposed evaluation metrics
for explanation tasks. 

\subsubsection{Fidelity}
First, the explanations should be faithful to the model. They should identify input features that are important for the model, not our humans. To evaluate this, the Fidelity+~\cite{pope2019explainability} metric is recently proposed. Intuitively, if important input features (nodes/edges/node features) identified by explanation techniques are discriminative to the model, the predictions should change significantly when these features are removed. 
Hence, Fidelity+ is defined as the difference of accuracy (or predicted probability) between the original predictions and the new predictions after masking out important input features~\cite{pope2019explainability, roar}.

Formally, let $\mathcal{G}_i$ denote the $i$-th input graph and $f(\cdot)$ denote the GNN classifier to be explained. The prediction result of this graph is represented as $\hat{y}_i = \mathbf{argmax}\ f(\mathcal{G}_i)$. Then its explanations can be considered as a hard importance map $m_i$ where each element is 0 or 1 to indicate if the corresponding feature is important. Note that for methods like ZORRO~\cite{anonymous2021hard} and Causal  Screening~\cite{anonymous2021causal}, the generated explanations are discrete masks, which can be directly used as the importance map. In addition, for methods like GNNExplainer~\cite{ying2019gnnexplainer} and GraphLime~\cite{huang2020graphlime}, the importance scores are continuous values, then the importance map $m_i$ can be obtained by normalization and thresholding or ranking. Finally, the Fidelity+ score of prediction accuracy can be computed as 
\begin{equation}
    Fidelity+^{acc} = \frac{1}{N}\sum_{i=1}^{N} (\mathbbm{1}(\hat{y}_i = y_i) - \mathbbm{1}(\hat{y}_i^{1-m_i} = y_i)),
\end{equation}
where $y_i$ is the original prediction of graph $i$ and $N$ is the number of graphs. Here $1-m_i$ means the complementary mask that removes the important input features and $\hat{y}_i^{1-m_i}$ is the prediction when feeding the new graph into trained GNN $f(\cdot)$. The indicator function $\mathbbm{1}(\hat{y}_i = y_i)$ returns 1 if $\hat{y}_i$ and $y_i$ are equal and returns 0 otherwise. Note that the $Fidelity+^{acc}$ metric studies the change of prediction accuracy. By focusing on the predicted probability, the Fidelity+  of probability can be defined as  
\begin{equation}
    Fidelity+^{prob} = \frac{1}{N}\sum_{i=1}^{N} (f(\mathcal{G}_i)_{y_i} - f(\mathcal{G}_i^{1-m_i})_{y_i}), 
\end{equation}
where  $\mathcal{G}^{1-m_i}_i$ represents the new graph obtained by keeping features of $\mathcal{G}_i$ based on the complementary mask $1-m_i$. Note that $Fidelity+^{prob}$ monitors the change of predicted probability, which is more sensitive than $Fidelity+^{acc}$. For both metrics, higher values indicate better explanation results and more discriminative features are identified.  

The Fidelity+ metric studies the prediction change by removing important nodes/edges/node features. In contrast, the metric Fidelity- studies prediction change by keeping important input features and removing unimportant features. Intuitively, important features should contain discriminative information so that they should lead to similar predictions as the original predictions even unimportant features are removed.
Formally, the metric Fidelity- can be computed as  

\begin{equation}
    Fidelity-^{acc} = \frac{1}{N}\sum_{i=1}^{N} (\mathbbm{1}(\hat{y}_i = y_i) - \mathbbm{1}(\hat{y}_i^{m_i} = y_i)),
\end{equation}
\begin{equation}
    Fidelity-^{prob} = \frac{1}{N}\sum_{i=1}^{N} (f(\mathcal{G}_i)_{y_i} - f(\mathcal{G}_i^{m_i})_{y_i}),
\end{equation}
where $\mathcal{G}^m_i$ is the new graph when keeping important features of $\mathcal{G}_i$ based on explanation $m_i$ and $\hat{y}_i^{m_i}$ is the new prediction. Note that for both $Fidelity-^{acc}$ and $Fidelity-^{prob}$, lower values indicate less importance information are removed so that the explanations results are better.

\subsubsection{Sparsity}

Second, good explanations should be sparse, which means they should capture the most important input features and ignore the irrelevant ones. The metric Sparsity measures such a property. Specifically, it measures the fraction of features selected as important by explanation methods~\cite{pope2019explainability}. Formally, give the graph $\mathcal{G}_i$ and its hard importance map $m_i$, the Sparsity metric can be computed as
\begin{equation}
    Sparsity = \frac{1}{N}\sum_{i=1}^{N}(1 - \frac{|m_i|}{|M_i|}),
\end{equation}
where $|m_i|$ denotes the number of important  input features (nodes/edges/node features) identified in $m_i$ and $|M_i|$ means the total number of features in $\mathcal{G}_i$. Note that higher values indicate the explanations are more sparse and tend to only capture the most important input information.

\subsubsection{Stability}

In addition, good explanations should be stable. Intuitively, when small changes are applied to the input without affecting the predictions, the explanations should remain similar. The recent proposed Stability metric measures whether an explanation method is stable~\cite{sanchez2020evaluating}. Given an input graph $\mathcal{G}_i$, its explanations $m_i$ is regarded as the ground truth. Then the input graph $\mathcal{G}_i$ is perturbed by small changes, such as attaching new nodes/edges, to obtain a new graph $\hat{\mathcal{G}_i}$. Note that $\mathcal{G}_i$ and $\hat{\mathcal{G}_i}$ are required to have the same predictions. Then the explanations of $\hat{\mathcal{G}_i}$ is obtained, denoted as $\hat{m_i}$. By comparing the difference between $m_i$ and $\hat{m_i}$, we can compute the Stability score. Note that lower values indicate the explanation technique is more stable and more robust to noisy information. In addition, since graph representations are sensitive, selecting a proper amount of perturbations may be challenging.

\subsubsection{Accuracy}

Furthermore, the Accuracy metric is proposed for synthetic datasets~\cite{sanchez2020evaluating, ying2019gnnexplainer}. In synthesis datasets, even though it is unknown whether the GNNs make predictions in our expected way, the rules of building these datasets, such as graph motifs, can be used as reasonable approximations of the ground truths. Then for any input graph, we can compare its explanations with such ground truths. For example, when studying important edges, we can study the matching rate for important edges in explanations compared with those in the ground truths.  The common metrics for such comparisons include general accuracy, F1 score, ROC-AUC score. Note that higher values indicate the explanations are closer to ground truths and can be considered as better results.  In addition, the Accuracy metric cannot be applied to real-world datasets due to the lack of ground truths.

\begin{figure*}[ht!]
    \centering
\includegraphics[width=2\columnwidth]{figure/fidelity.png}
    \caption{The Fidelity+ comparisons between different GNN explanation techniques under different Sparsity levels. }
    \label{fig:Fidelity+}
\end{figure*}

\begin{figure*}[ht!]
    \centering
\includegraphics[width=2\columnwidth]{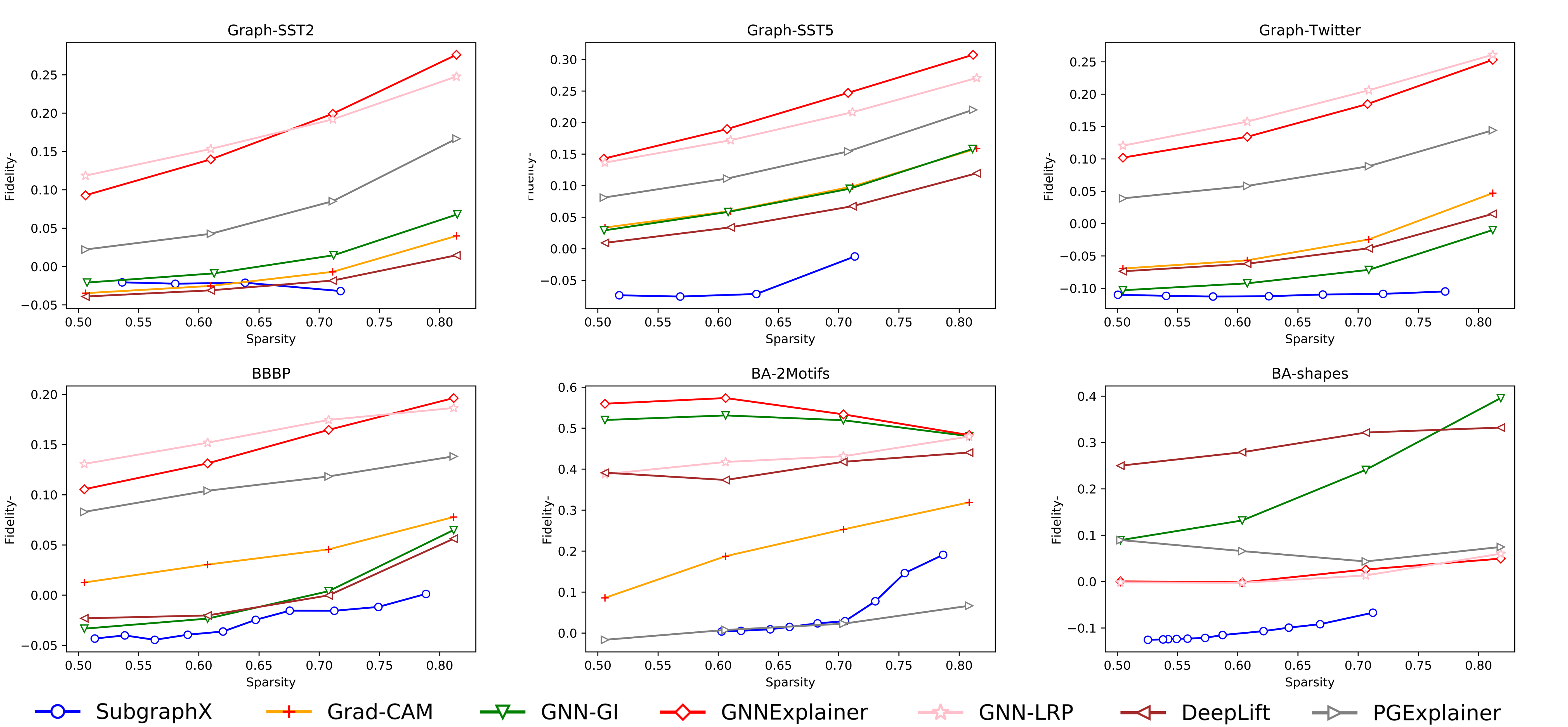}
    \caption{The Fidelity- comparisons between different GNN explanation techniques under different Sparsity levels.}
    \label{fig:Fidelity-}
\end{figure*}

\subsubsection{Discussions}

It is noteworthy that different metrics should be combined to evaluate explanation results. For example, Sparsity and Fidelity+/Fidelity- are highly correlated. When the explanation results are soft values, the Sparsity is determined by a threshold value. Intuitively, larger threshold values tend to identify fewer features as important and hence increase the Sparsity score and decrease the Fidelity+ score. Hence, to fairly compare different explanation methods, we suggest comparing their Fidelity+ scores with the same level of Sparsity scores. One possible way is to select a fixed percentage of input features as important, and then compare their Fidelity+ or Fidelity- scores.     

In addition, there are several other metrics to evaluate explanation results, such as Contrastivity~\cite{pope2019explainability} and Consistency~\cite{sanchez2020evaluating}. The Contrastivity score is based on a strong assumption that the explanations for different classes should be significantly different, which ignores the common patterns among different classes. Meanwhile, the Consistency metric assumes that high-performing model architectures should have consistent explanations. However, different models may capture different relationships even though they achieve competitive performance, especially for large-scale and complex datasets. Hence, we believe such metrics may not be fair enough to evaluate explanation results and they are not listed as recommended metrics.

\newcolor{\section{Experimental Studies}}
To better understand different explanation techniques, we conduct experiments to compare different methods and study our evaluation metrics. \newcolorthree{We provide a benchmark that can be used to reproduce our reported results and conduct experiments to compare different techniques for future studies\footnote{\newcolorthree{\url{https://github.com/divelab/DIG/tree/dig/benchmarks/xgraph}}}.}

\begin{table*}[]
\caption{The Fidelity+ comparisons between different GNN explanation techniques and the random designation baseline. }
\centering
\begin{tabular}{lcccccc}
\hline
Methods      & Graph-Twitter & Graph-SST2 & Graph-SST5 & BBBP                          & BA-2Motifs                     & BA-Shapes        \\ \hline
             & \multicolumn{5}{c}{Sparsity=0.7}                                                                        & Sparsity=0.6                   \\ \hline
Random       & 0.1342        & 0.0915    & 0.1419     & 0.1212                        & 0.4903                         & 0.1884                         \\ \hline
SubgraphX    & 0.2836        & 0.3152     & 0.2351     & 0.4521                        & 0.8642                         & 0.3171                         \\ \hline
Grad-CAM     & 0.2418        & 0.2414     & 0.2118     & 0.2036                        & 0.6112                         & N/A                            \\ \hline
GNN-GI       & 0.2593        & 0.2571     & 0.2031     & 0.3051                        &  0.0466* &  0.1723*  \\ \hline
GNNExplainer & 0.1452        & 0.0953     & 0.1441     &  0.1057* &  0.4972  & 0.2925                         \\ \hline
PGExplainer  & 0.1704        & 0.1889     & 0.1854     & 0.1464                        &  0.1126*  & 0.2015                         \\ \hline
GNN-LRP      & 0.1931        & 0.1363     & 0.1813     &  0.0860*  &  0.5125  & 0.3386                        \\ \hline
DeepLift     & 0.2336        & 0.2454     & 0.1924     & 0.3039                        &  0.2156*  &  0.0411* \\ \hline
\end{tabular}
\label{table:fidelity}
\end{table*}

\begin{table*}[]
\caption{The Fidelity- comparisons between different GNN explanation techniques and the random designation baseline. }
\centering
\begin{tabular}{lcccccc}
\hline
Methods      & Graph-Twitter & Graph-SST2 & Graph-SST5 & BBBP     & BA-2Motifs & BA-Shapes      \\ \hline
             & \multicolumn{5}{c}{Sparsity=0.7}                               & Sparsity=0.6                 \\ \hline
Random       & 0.2825        & 0.2745     & 0.2961     & 0.2168   & 0.5394     & 0.2567                       \\ \hline
SubgraphX    & -0.1085       & -0.0288  & -0.0298   & -0.0169 & 0.0686   & -0.0792                      \\ \hline
Grad-CAM     & -0.0245       & -0.0069    & 0.0987     & 0.0456  & 0.2529     & N/A                         \\ \hline
GNN-GI       & -0.0715       & 0.0147     & 0.0951     & 0.0039 & 0.5193     & 0.1318                       \\ \hline
GNNExplainer & 0.1848        & 0.1992     & 0.2471     & 0.1647   & 0.5337     & -0.0017                   \\ \hline
PGExplainer  & 0.0887        & 0.0852     & 0.1543     & 0.1183   & 0.0227    & 0.0658                      \\ \hline
GNN-LRP      & 0.2060         & 0.1919     & 0.2164     & 0.1746   & 0.4314     & -0.0026                   \\ \hline
DeepLift     & -0.0382       & -0.0183    & 0.0674     & -0.0002 & 0.4179     & 0.2790* \\ \hline
\end{tabular}
\label{table:fidelity-}
\end{table*}

\begin{table*}[]
\caption{\newcolortwo{The Accuracy and Stability comparisons between different GNN explanation techniques.}}
\centering
\begin{tabular}{lcccc}
\hline
Methods      & \multicolumn{2}{c}{BA-shapes}  & \multicolumn{2}{c}{BA-Community}  \\ \hline
Metric       & Accuracy      & Stability    & Accuracy      & Stability     \\  \hline 
GNN-GI       & 0.8369        & 0.1361       & 0.8291        & 0.1723        \\ \hline
GNNExplainer & 0.8786        & 0.1721       & 0.9194        & 0.1820         \\ \hline
PGExplainer  & 0.7147        & 0.0522       & 0.6843        & 0.1177        \\ \hline
GNN-LRP      & 0.9243        & 0.1872       & 0.8357        & 0.1239        \\ \hline
DeepLift     & 0.5698        & 0.0432      & 0.4190         & 0.0842      \\ \hline
\end{tabular}
\label{table:accuracy_stablebility}
\end{table*}

\newcolor{\subsection{Method Comparisons with Fidelity}}
\newcolor{We first compare several GNN explanation techniques using the Fidelity and Sparsity metrics. Specifically, we compare the Fidelity scores under different Sparsity levels. Note that here we employ the probability version $Fidelity+^{prob}$ and $Fidelity-^{prob}$ as the example. To provide a comprehensive study, we employ six different datasets; these are, Graph-SST2, Graph-SST5, Graph-Twitter, BBBP, BA-2Motifs, and BA-Shapes, which include both graph classification and node classification tasks, both real-world and synthetic data. In addition, we use GNNs as the graph model to compare different techniques. The results are reported in Figure~\ref{fig:Fidelity+} and~\ref{fig:Fidelity-}. Note that GNN-GI is a simplified version of GNN-LRP, which is developed based on the $\mbox{gradient}\times \mbox{input}$ rule. DeepLift~\cite{shrikumar2017learning} is a popular decomposition method in the computer vision domain and we also extend it to explain deep graph models.} \newcolortwo{In addition, since Grad-CAM cannot be directly applied to node classification model, we ignore it for the BA-Shapes dataset.}

In Figure~\ref{fig:Fidelity+}, we focus on the Fidelity+ metric, which measures the prediction probability drop when important features are removed. Higher Fidelity+ scores indicate the identified features are more important for the GNN model. Obviously, SubgraphX outperforms the other methods significantly and consistently, showing that subgraph-level explanations are more desired for GNN models. In addition, the flow-based techniques, GNN-GI and GNN-LRP, generally obtain higher Fideity+ scores than GNNExplainer and PGExplainer. Based on these facts, we believe the explanations considering structural information are more suitable for explaining GNN models. An interesting finding is that GNNExplainer and PGExplainer perform well on the node classification task (BA-Shapes) but not as expected for the graph classification models. 
We also observe that the performance of traditional methods, DeepLift and Grad-CAM, are not very stable across different datasets. \newcolortwo{We believe the main reason is that these two methods are directly migrated from the image domain and lack graph specific designs. }

\newcolor{In Figure~\ref{fig:Fidelity-}, we study the Fidelity- metric, which measures the prediction probability drop when important features are kept and unimportant features are removed. Lower values mean that fewer important features are removed so that the explanations are better. The observations are similar to the results of Fidelity+. SubgraphX still obtains stable and the best results while flow-based techniques, GNN-GI and GNN-LRP, are performing better than the others. In addition, GNNExplainer and PGExplainer still only perform well on the node classification task.}

\begin{figure*}[ht!]
    \centering
\includegraphics[width=2\columnwidth]{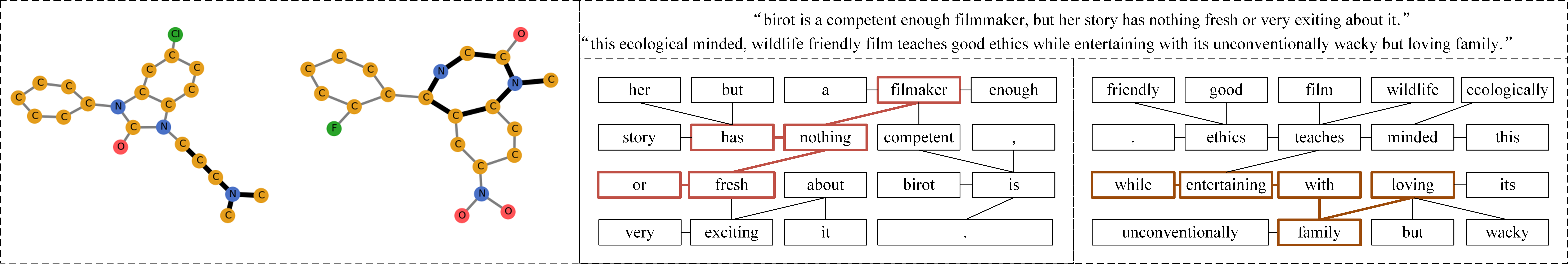}
    \caption{\newcolortwo{Visualizations of explanations on the BBBP and Graph-SST2 datasets.}}
    \label{fig:vis_mol_nlp}
\end{figure*}
\newcolor{From our observations, we believe it is necessary to consider graph structures when developing GNN explanation techniques. In addition, when selecting baselines for future studies, we would recommend that select SubgraphX for all cases, use GNNExplainer and PGExplainer only for node classification datasets, and choose GNN-GI and GNN-LRP only for real-world datasets.} 

\newcolor{\subsection{Study of Fidelity}}
\newcolor{We further study our proposed evaluation metric Fidelity with a random designation of feature importance~\cite{roar}. Specifically, we assign the random importance scores to different edges and compare the performance of different methods to this random baseline. It helps answer two questions; these are, is a particular GNN explanation method better than a random guess? and can the evaluation metrics successfully capture the difference between them? Intuitively, good evaluation metrics should demonstrate the performance difference between well-performed models and the random baseline. The results are reported in Table~\ref{table:fidelity} and~\ref{table:fidelity-}. }

\begin{figure*}[ht!]
    \centering
\includegraphics[width=1.8\columnwidth]{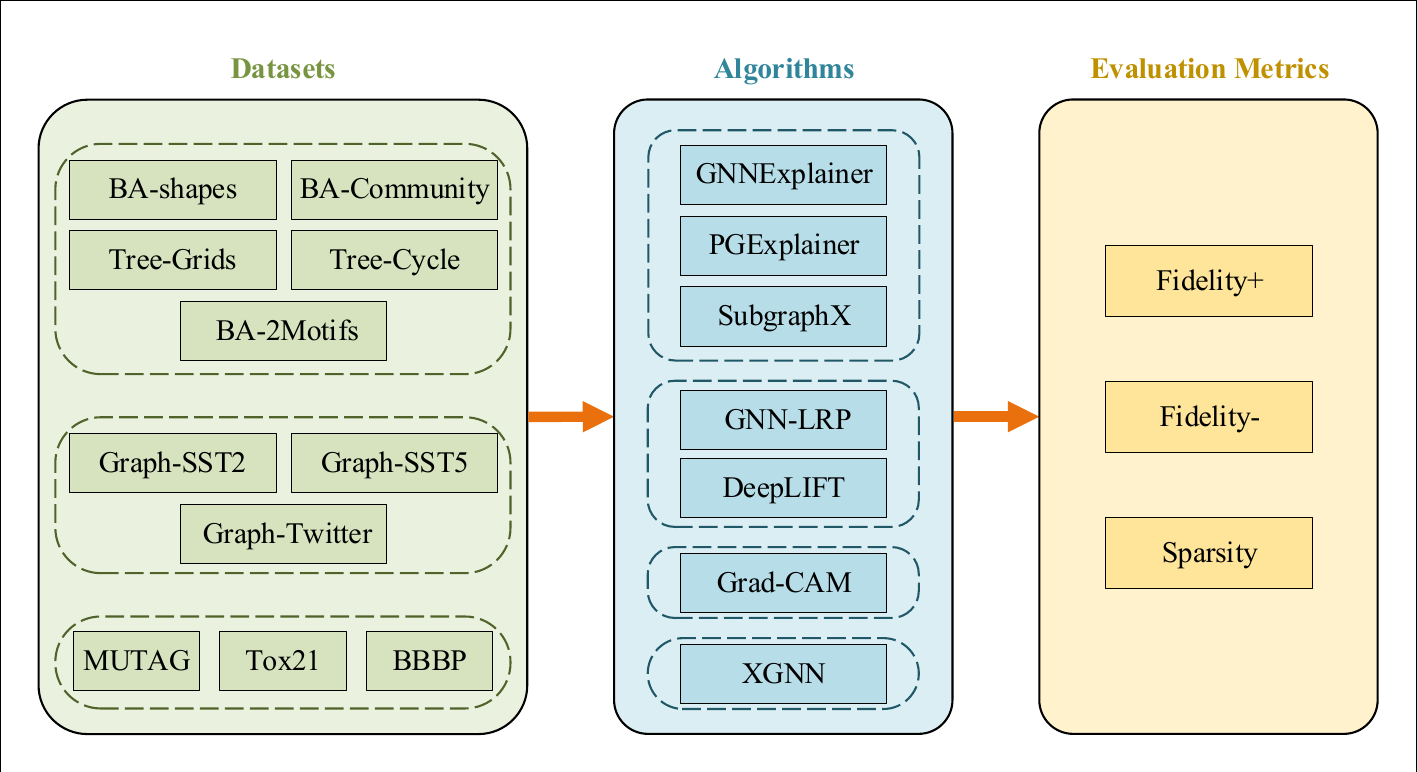}
    \caption{A overview of our open-source library, which includes unified implementations of multiple common baseline methods, commonly used datasets, and three evaluation metrics.}
    \label{fig:library}
\end{figure*}

\newcolor{We show the Fidelity scores of different techniques with Sparsity levels equal to 0.6 and 0.7. Note that the results that are inferior to the random baseline are denoted with $*$. We can observe that the Fidelity metrics show that different GNN explanation techniques generally obtain better results than the random guess. Such results indicate the correctness of our proposed Fidelity metrics.} \newcolortwo{Meanwhile, we notice that
the results of DeepLift are worse than the random baseline on BA-Shapes and BA-Motifs datasets. We believe the reason is that DeepLift does not consider graph information and hence is not stable across datasets, which is consistent with our findings from Figure~\ref{fig:Fidelity+} and \ref{fig:Fidelity-}. In addition, GNN-GI is also not performing well on these two datasets. One possible reason is GNN-GI is developed based on the $\mbox{gradient}\times \mbox{input}$ rule so its results are highly depending on the properties of input graphs.
Furthermore, all methods are performing better than the random baseline for both Fidelity+ and Fidelity- metrics on our proposed sentiment graph datasets. It indicates that our proposed datasets are suitable and stable to evaluate different GNN explanation techniques.}

\newcolortwo{\subsection{Method Comparisons with Accuracy}}
\newcolortwo{In this section, we further compare different techniques using the Accuracy metric. Since the Accuracy metric requires the ground truths for GNN explanations, we apply extensive experiments on synthetic datasets.
We evaluate different methods using the BA-shapes and BA-Community datasets, which are taken from 
GNNExplainer~\cite{ying2019gnnexplainer}. Specifically, the ground-truth motifs are converted to ground-truth edge masks and we compute the AUC-ROC score between the edge importance scores provided by different explanation techniques and the ground-truth edge masks. Note that SubgraphX provides subgraph-level explanations, which cannot be converted to soft edge masks. Hence, SubgraphX is not directly comparable in terms of AUC-ROC score and is ignored. In addition, since these two datasets are node classification tasks, Grad-CAM is also ignored. The results are reported in Table~\ref{table:accuracy_stablebility}. We observe that GNN-LRP obtains the highest Accuracy on the BA-shapes dataset and GNNExplainer performs the best on the BA-Community dataset. Overall, we believe GNN-LRP, GNNExplainer, and GNN-GI have competitive performance while PGExplainer is not performing as promising as its original reported results. Our finding regarding PGExplainer is similar to a recent reproducibility work~\cite{holdijk2021re}. In addition, DeepLift performs the worst on both datasets, which is reasonable because DeepLift does not consider any graph-specific information when providing explanations. Finally, it is noteworthy that the ground truths in synthetic datasets are only reasonable approximations of
the ground truths instead of real ground truths.} 

\newcolortwo{\subsection{Method Comparisons with Stability}}
\newcolortwo{We also compare different methods using the Stability metric. The key idea of Stability is to apply small changes to the input graph without affecting the predictions and measure the difference between explanations. We follow an existing work~\cite{sanchez2020evaluating} to set up the experiments on BA-shapes and BA-Community datasets. Specifically, for each node, we randomly attach two noise nodes to its L-hop neighbor graph while keeping the ground-truth motifs unchanged. Then we generate 10 input perturbations for each node and obtain the corresponding explanations. For each pair of perturbation and explanation, we evaluate it using the Accuracy metric above. Finally, the Stability score is the absolute value of the averaged Accuracy score change for these 10 input perturbations compared to the original graph. The results are reported in Table~\ref{table:accuracy_stablebility}. Note that since Stability is developed based on Accuracy and these two datasets are node classification datasets, SubgraphX and Grad-CAM are not included. We notice that DeepLift has the best Stability score but its corresponding Accuracy scores are significantly worse. Such results indicate that DeepLift performs not well consistently and stably. We also observe that the method with higher Accuracy tends to have a higher Stability score (note that a lower Stability score means more stable). Then it is important to analyze the results considering both of them and a possible way to understand the Stability score is to consider it as the error bar of the Accuracy score. Hence, for those three methods with competitive Accuracy scores, we believe GNN-LRP and GNNExplainer have better performance than GNN-GI in terms of Accuracy and Stability. 
}

\newcolortwo{\subsection{Dataset Visualizations}}
\newcolortwo{In this section, we present the visualized results from molecular datasets and our proposed graph sentiment datasets. The results are reported in Figure~\ref{fig:vis_mol_nlp}. The left two examples are the visualized explanations from the BBBP~\cite{martins2012bayesian} dataset and both examples are labeled and predicted as positive. Note that a positive label means the molecule has the Blood-Brain-Barrier Penetration property. While the CN chain is identified in the first example and the CN ring is captured in the second example, it is challenging for the users to understand the explanations without chemical domain knowledge. It's not clear what is the chemical meaning of these patterns. Hence, we believe such results are less intuitive and human-intelligible for non-experts. The third and fourth examples are from the Graph-SST2 dataset and both are correctly predicted. The third one is negative and the last one is positive. Since each node is associated with a word or a symbol, users can understand the sentiment meaning of each node. Then it is easier to understand the identified explanations.
For example, in the third example, the phrase ``has nothing fresh'' has a clearly negative sentiment. 
Hence, we believe our proposed graph sentiment datasets are more suitable for evaluating GNN explanation techniques.}

\section{An Open-Source Library}
\newcolor{To facilitate research into graph neural networks, we develop an open-source library to study the explainability in GNNs. It is developed as a part of our DIG (Dive into Graphs) library~\cite{liu2021dig}, which is a turnkey library for graph deep learning research and now 
publicly available\footnote{\url{https://github.com/divelab/DIG/}}.} Our library can be directly used to reproduce results of existing GNN explanation methods, develop new algorithms, and conduct evaluations for explanation results. First, in this library, we include the implementations of several existing methods, such as GNNExplainer~\cite{ying2019gnnexplainer}, PGExplainer~\cite{luoparameterized}, DeepLIFT~\cite{shrikumar2017learning}, GNN-LRP~\cite{schnake2020higherorder}, Grad-CAM~\cite{pope2019explainability}, SubgraphX~\cite{yuan2021explainability}, and XGNN~\cite{xgnn}. While several methods provide official implementations, their development environments may vary, thus making fair comparisons difficult.

To this end,  we implement these methods in the same development environment that the key dependencies of our library are PyTorch 1.6.0~\cite{NEURIPS2019_9015} and PyTorch Geometric 1.6.1~\cite{Fey2019Geometric}. With such a unified environment, new techniques can be easily and fairly compared with existing methods. Second, we include several datasets for GNN explanation tasks in our library. Note that we not only provide the aforementioned sentiment graph datasets, but also include several commonly used datasets, such as MUTAG, BBBP, BA-shapes, and BA-2Motifs, etc.  These datasets
include synthetic data, text data, and molecular data, and cover both graph classification tasks and node classification tasks. Hence, we believe that using such datasets can provide a comprehensive study of different GNN explanation methods. In addition, our library provides implementations of different evaluation metrics, including Fidelity+, Fidelity-, and Sparsity. These metrics can be directly applied to evaluate the explanation results obtained from our framework and provide comprehensive evaluations from the model's perspective.

\section{Conclusion}
Graph Neural Networks are widely studied recently but the explainability of graph models is still less explored. To investigate the underlying mechanisms of these black-boxes, several explanation methods are proposed to explain graph models, including XGNN, GNNExplainer, etc. These methods explain graph models from different views and motivations, thus lacking a comprehensive study and analysis of these methods. In this work, we provide a systematic and comprehensive survey for these methods. We first propose a novel taxonomy of existing GNN explanation methods and introduce the key idea behind each category. Then we discuss each explanation method in detail, including the methodology, insights, advantages, and drawbacks. In addition, we provide a  comprehensive analysis of different explanation methods. Furthermore, we introduce and analyze the commonly employed datasets and metrics for evaluating GNN explanation methods. Last but not the least, we build three graph datasets from text data, which are human-understandable and can be directly used for GNN explanation tasks.


%

\ifCLASSOPTIONcompsoc
  \section*{Acknowledgments}
\else
  \section*{Acknowledgment}
\fi

This work was supported in part by National Science Foundation grants IIS-1955189 and DBI-2028361.

\ifCLASSOPTIONcaptionsoff
  \newpage
\fi



%
\bibliographystyle{IEEEtran}
\bibliography{xgraph}

\end{document}